\documentclass[10pt,twocolumn,letterpaper]{article}

\usepackage[pagenumbers]{cvpr} 

\usepackage{graphicx}
\usepackage{amsmath}
\usepackage{amssymb}
\usepackage{booktabs}
\usepackage{bm}
\usepackage{adjustbox}

\usepackage[pagebackref,breaklinks,colorlinks]{hyperref}

\usepackage[capitalize]{cleveref}
\crefname{section}{Sec.}{Secs.}
\Crefname{section}{Section}{Sections}
\Crefname{table}{Table}{Tables}
\crefname{table}{Tab.}{Tabs.}

\def\confName{CVPR}
\def\confYear{2022}

\begin{document}

\title{Semantic-Aware Implicit Neural Audio-Driven Video Portrait Generation}

\author{Xian Liu$^{1}$, Yinghao Xu$^{1}$, Qianyi Wu$^{2}$, Hang Zhou$^{1}$, Wayne Wu$^{3}$, Bolei Zhou$^{1}$\\
    
    $^1$The Chinese University of Hong Kong \quad $^2$Monash University \quad
    $^3$SenseTime Research\\

    {\tt\small \{alvinliu@ie, xy119@ie, zhouhang@link, bzhou@ie\}.cuhk.edu.hk, }\\ 
    {\tt\small qianyi.wu@monash.edu, wuwenyan@sensetime.com}
}

\maketitle


\begin{abstract}
Animating high-fidelity video portrait with speech audio is crucial for virtual reality and digital entertainment. While most previous studies rely on accurate explicit structural information, recent works explore the implicit scene representation of Neural Radiance Fields (NeRF) for realistic generation. In order to capture the inconsistent motions as well as the semantic difference between human head and torso, some work models them via two individual sets of NeRF, leading to unnatural results. In this work, we propose Semantic-aware Speaking Portrait NeRF (SSP-NeRF), which creates delicate audio-driven portraits using one unified set of NeRF. The proposed model can handle the detailed local facial semantics and the global head-torso relationship through two semantic-aware modules. Specifically, we first propose a Semantic-Aware Dynamic Ray Sampling module with an additional parsing branch that facilitates audio-driven volume rendering. Moreover, to enable portrait rendering in one unified neural radiance field, a Torso Deformation module is designed to stabilize the large-scale non-rigid torso motions. Extensive evaluations demonstrate that our proposed approach renders more realistic video portraits compared to previous methods. Project page: \href{https://alvinliu0.github.io/projects/SSP-NeRF}{https://alvinliu0.github.io/projects/SSP-NeRF}.

\end{abstract}
\section{Introduction}
\label{sec:1}
Generating high-fidelity video portraits based on speech audio is of great importance to various applications like digital human, film-making and video dubbing. Many researchers tackle the task of audio-driven talking face or video portrait generation by using deep generative models. 
Several works rely solely on learning-based image reconstruction, which typically synthesize static results of low-resolution ~\cite{chen2018lip, chung2017you, zhou2021pose, zhou2019talking, vougioukas2020realistic, prajwal2020lip, wiles2018x2face}. Other methods utilize explicit structural intermediate representations such as 2D landmarks~\cite{chen2019hierarchical, das2020speech, suwajanakorn2017synthesizing} or 3D facial models~\cite{chen2020talking, richard2021audio, thies2020neural, zhou2020makelttalk, yi2020audio, wang2021one, wu2021imitating}. Though some of them can generate high-fidelity images~\cite{thies2020neural,suwajanakorn2017synthesizing}, the errors in structured representation prediction (\emph{e.g.}, expression parameters of a 3D Morphable Model (3DMM)~\cite{blanz1999morphable}) lead to inaccurate face deformation~\cite{zhou2021pose}.

Recently, the implicit 3D scene representation of Neural Radiance Fields (NeRF)~\cite{mildenhall2020nerf} provides a new perspective for realistic generation. It enables free-view control with higher image quality compared to explicit methods, which is suitable for the video portrait generation task. Gafni \textit{et al.}~\cite{Gafni_2021_CVPR} first involve NeRF in the dynamic human head modeling from single-view data in a video-driven manner. However, an accurate explicit 3D model is still required in their settings. Moreover, they model torso consistently with the head, which leads to unstable results. Guo \textit{et al.}~\cite{guo2021adnerf} further propose AD-NeRF for audio-driven talking head synthesis. In particular, they build two individual sets of NeRF for head and torso modeling conditioned on audio input. Such a straightforward pipeline suffers from head-torso separation during the render stage, making generated results unnatural.

Based on previous studies, we identify two key challenges for incorporating NeRF into video portrait generation: 1) Each facial part's appearance and moving patterns are intrinsically connected but substantially different, especially when associated with audios. Thus weighing all rendering areas equally without semantic guidance would lead to blurry details and difficulties in training. 2) While it is easy to bind head pose with camera pose, the global movements of the head and torso are in significant divergence. As the human head and torso are non-rigidly connected, modeling them with one set of NeRF is an ill-posed problem.

In this work, we develop a method called Semantic-aware Speaking Portrait NeRF (\textbf{SSP-NeRF}), which generates stable audio-driven video portraits of high-fidelity. We show that \emph{semantic awareness is the key to handle both local facial dynamics and global head-torso relationship}. Our intuition lies in the fact that different parts of a speaking portrait have different associations with speech audio. While other organs like ears move along with the head, the high-frequency mouth motion that is strongly correlated with audios requires additional attention. To this end, we devise an \emph{Semantic-Aware Dynamic Ray Sampling} module, which consists of an \emph{Implicit Portrait Parsing} branch and a \emph{Dynamic Sampling Strategy}. Specifically, the parsing branch supervises the modeling with facial semantics in 2D plane. Then the number of rays sampled at each semantic region could be adjusted dynamically according to the parsing difficulty. Thus more attention can be paid to the small but important areas like lip and teeth for better lip-synced results. Besides, we also enhance the semantic information by anchoring a set of latent codes to the vertices of a roughly predicted 3DMM~\cite{blanz1999morphable} without expression parameters. 

On the other hand, since the head and torso motions are rigidly bound together in the current NeRF, a correctly positioned torso cannot be rendered even with the portrait parsing results. We further observe the relationship between head and torso: while they share the same translational movements, the orientation of torso seldom changes with head pose under the speaking portrait setting. Thus we model non-rigid deformation through a \emph{Torso Deformation} module. Concretely, for each point $(x, y, z)$ in the 3D scene, we predict a displacement $(\Delta x, \Delta y, \Delta z)$ based on the head-canonical view information and time flows. Interestingly, although there are local deformations on the face, the deformation module implicitly learns to focus on the global parts. This design facilitates portrait stabilization in one unified set of NeRF. Experiments demonstrate that our method generates high-fidelity video portraits with better lip-synchronization and better image quality efficiently.

To summarize, our work has three main contributions: \textbf{(1)} We propose the \emph{Semantic-Aware Dynamic Ray Sampling} module to grasp the detailed appearance and local dynamics of each portrait part without using accurate structural information. \textbf{(2)} We propose the \emph{Torso Deformation} module that implicitly learns the global torso motion to prevent unnatural head-torso separated results. \textbf{(3)} Extensive experiments show that the proposed \textbf{SSP-NeRF} renders high-fidelity audio-driven video portraits with one unified NeRF in an efficient manner, which outperforms state-of-the-art methods on both objective evaluations and human studies.
\section{Related Work}
\noindent\textbf{Audio-Driven Talking Head Synthesis.} Driving talking head with speech audio has a bunch of applications, which is of great research interest to computer vision and graphics. Conventional works mostly resort to stitching techniques~\cite{fisher1968confusions, brand1999voice, bregler1997video}, where a predefined set of phoneme-mouth correspondence rules is used to modify mouth shapes. With the rapid growth of deep neural networks, end-to-end frameworks are proposed. One category of methods, namely image reconstruction-based methods, generate talking face by latent feature learning and image reconstruction~\cite{chung2017you, zhu2018arbitrary, zhou2021pose, zhou2019talking, vougioukas2020realistic, prajwal2020lip, pham2017speech, taylor2017deep, cudeiro2019capture, zhang2021flow, wang2021audio2head, song2018talking,ji2021audio-driven}. For example, Chung \etal~\cite{chung2017you} propose the first end-to-end method with an encoder-decoder pipeline. Zhou \etal~\cite{zhou2019talking} explicitly disentangle identity and word information for better feature extraction. Prajwal \etal~\cite{prajwal2020lip} achieve synchronous lip movements with a pretrained lip-sync expert. However, these methods can only generate fix-sized images with low resolution. Another line of approaches named model-based methods utilize structural intermediate representations like 2D facial landmarks or 3D representations to bridge the mappings from audio to complicated facial images~\cite{chen2019hierarchical, das2020speech, suwajanakorn2017synthesizing, wang2020mead, chen2020talking, song2020everybody, thies2020neural, zhou2020makelttalk, yi2020audio, wu2021imitating, meshry2021learned}. Typically, Chen \etal~\cite{chen2019hierarchical} and Das \etal~\cite{das2020speech} first predict 2D landmarks then generate faces. Thies \etal~\cite{thies2020neural} and Song \etal~\cite{song2020everybody} infer facial expression parameters from audio in the first stage, then generate 3D mesh for final image synthesis. But errors in intermediate prediction often hinder accurate results. In contrast to these two lines of works, our method can render more realistic speaking portraits of high-fidelity without any accurate structural information.

\noindent\textbf{Implicit Representation Methods.} Recent works leverage implicit functions for learning scene representations~\cite{sitzmann2019scene, mildenhall2020nerf, liu2020dist, niemeyer2020differentiable, liu2020neural, zhang2021neural}, where multi-layer perceptron (MLP) weights are used to represent the mapping from spatial coordinates to a signal in continuous space like occupancy~\cite{mescheder2019occupancy, peng2020convolutional, saito2019pifu, ren2021csg}, signed distance function~\cite{yariv2020multiview, wang2021neus, gropp2020implicit}, color and volume density~\cite{mildenhall2020nerf, martin2021nerf, barron2021mipnerf, deng2021depth}, semantic label~\cite{Zhi:etal:ICCV2021, kohli2020inferring} and neural feature map~\cite{chen2021learning, niemeyer2021giraffe}. A recent popular work named Neural Radiance Fields (NeRF)~\cite{mildenhall2020nerf} optimizes an underlying continuous volumetric scene mapping from 5D coordinate of spatial location and view direction to implicit fields of color and density for photo-realistic view results. Naturally, naive NeRF is confined to static scenes, which triggers a branch of studies to extend NeRF for dynamic scenes~\cite{park2021nerfies, pumarola2021d, tretschk2021non, peng2021neural, peng2021animatable, noguchi2021neural, Gafni_2021_CVPR, raj2021pva, park2021hypernerf, liu2021neural, sun2021nelf, palafox2021neural, bozic2021neural, meka2020deep}. However, few works focus on complicated dynamic scenes like speaking portraits. The main difficulty lies in the learning of cross-modal associations between different portrait parts and speech audio. One relevant work~\cite{guo2021adnerf} synthesizes talking head with two individual sets of NeRF for head and torso, making generated results fall apart. In this work, we take semantics as guidance to grasp each portrait part's local dynamics and appearances for fine-grained results efficiently. A deformation module further enables us to synthesize stable video portraits using one unified set of NeRF.
\section{Our Approach}
\label{sec:3}
\begin{figure*}[t!]
    \centering
    \includegraphics[width=1\linewidth]{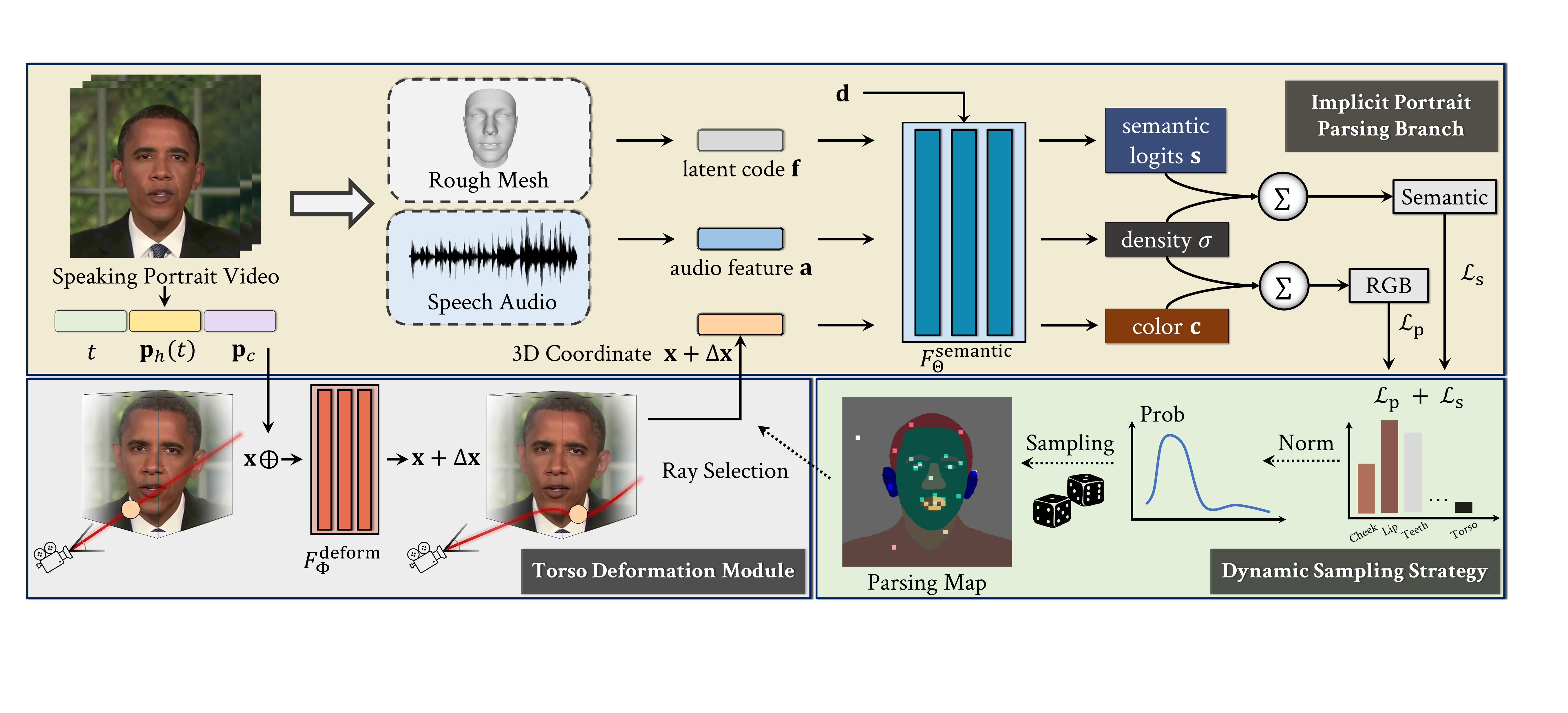}
    \caption{\textbf{Overview of Semantic-aware Speaking Portrait NeRF (SSP-NeRF) framework.} In Implicit Portrait Parsing Branch (\textcolor[rgb]{0.9, 0.751, 0.4}{yellow}), the semantic-aware implicit function $F^{\text{semantic}}_{\Theta}$ takes latent code $\mathbf{f}$, audio feature $\mathbf{a}$, 3D coordinate $\mathbf{x}$ and view direction $\mathbf{d}$ as input, then outputs the semantic logits $\mathbf{s}$, density $\sigma$ and color $\mathbf{c}$ of the scene. In Dynamic Sampling Strategy (\textcolor[rgb]{0.32941176470588, 0.6, 0.2078431372549}{
   green}), the RGB loss $\mathcal{L}_\mathrm{p}$ and semantic loss $\mathcal{L}_\mathrm{s}$ are utilized to guide the distribution of rays sampled at each semantic region. In particular, the Torso Deformation module (\textcolor[rgb]{0.37647058823529, 0.37647058823529, 0.37647058823529}{grey}) uses an implicit function $F^{\text{deform}}_{\Phi}$ to map from the time $t$, head pose $\mathbf{p}_\mathrm{h}(t)$, canonical pose $\mathbf{p}_\mathrm{c}$ and 3D coordinate $\mathbf{x}$ into the displacement $\Delta\mathbf{x}$, which generates the deformed 3D coordinate $\mathbf{x} + \Delta\mathbf{x}$ to model non-rigid torso motions.}
    \label{fig:framework}
    \vspace{-1mm}
\end{figure*}

We present \textbf{Semantic-aware Speaking Portrait NeRF (SSP-NeRF)} that generates delicate audio-driven portraits with one unified set of NeRF. The whole pipeline is depicted in Fig.~\ref{fig:framework}. In this section, we first review the preliminaries and the problem setting of video portrait synthesis with neural radiance fields (Sec.~\ref{sec:3.1}). We then introduce the \emph{Semantic-Aware Dynamic Ray Sampling} module, 
which facilitates fine-grained appearance and dynamics modeling for each portrait part with semantic information (Sec.~\ref{sec:3.2}). Furthermore, we elaborate the \emph{Torso Deformation} module that handles non-rigid torso motion by learning location displacements (Sec.~\ref{sec:3.3}). Finally, the volume rendering process and network training details are described (Sec.~\ref{sec:3.4}).

\subsection{Preliminaries and Problem Setting}
\label{sec:3.1}
Given images with calibrated camera intrinsics and extrinsics, NeRF~\cite{mildenhall2020nerf} represents a scene using a continuous volumetric radiance field $F$. Specifically, $F$ is modeled by an MLP, which takes 3D spatial coordinates $\mathbf{x} = (x, y, z)$ and 2D view directions $\mathbf{d} = (\theta, \phi)$ as input, then outputs the implicit fields of color $\mathbf{c} = (r, g, b)$ and density $\sigma$. In this way, the MLP weights store scene information by the mapping of $F : (\mathbf{x}, \mathbf{d}) \rightarrow (\mathbf{c}, \sigma)$. To compute the color of a single pixel, NeRF~\cite{mildenhall2020nerf} approximates the volume rendering integral using numerical quadrature~\cite{max1995optical}. Consider the ray $\mathbf{r}(v) = \mathbf{o} + v\mathbf{d}$ from camera center $\mathbf{o}$, its expected color $\hat{C}(\mathbf{r})$ with near and far bounds $v_n$ and $v_f$ is calculated as:
\begin{equation} 
    \label{eq:nerf}
\hat{C}(\mathbf{r}) = \int_{v_n}^{v_f}T(v)\sigma(\mathbf{r}(v))\mathbf{c}(\mathbf{r}(v), \mathbf{d})dv,
\end{equation}
where $T(v) = \exp(-\int_{v_n}^v\sigma(\mathbf{r}(u))du)$ is the accumulated transmittance along the ray from $v_n$ to $v$. With the hierarchical volume sampling, both coarse and fine MLPs are optimized by minimizing the photometric discrepancy.

In this work, we focus on audio-driven video portrait generation in a basic setting: 1) The camera pose $\{R, \tau\}$ is given by the estimated rigid head pose, where the rotation matrix $R \in \mathbb{R}^{3 \times 3}$ and the translation vector $\tau \in \mathbb{R}^{3 \times 1}$ are estimated by 3DMM~\cite{blanz1999morphable} on the face; 2) The audio feature $\mathbf{a} \in \mathbb{R}^{64}$ is extracted by a pretrained DeepSpeech~\cite{amodei2016deep} model and further processed with a light-weight audio encoder to get more compact representation. Therefore, the implicit function of audio-driven portrait \textbf{basic} setting is:
\begin{equation} 
    \label{eq:basic}
F^{\text{basic}} : (\mathbf{x}, \mathbf{d}, \mathbf{a}) \rightarrow (\mathbf{c}, \sigma).
\end{equation}

Guo \etal~\cite{guo2021adnerf} use an off-the-shelf parsing method~\cite{lee2020maskgan} to divide training images into head and torso for individual NeRF modeling. Following their settings, we assume that the semantic parsing maps are also available in our method.
\subsection{Semantic-Aware Dynamic Ray Sampling}
\label{sec:3.2}
To avoid the unnatural head-torso separation problem described in Sec.\ref{sec:1}, we render the whole portrait with one unified set of NeRF. However, two problems remain: 1) The associations between different portrait parts and audio are different. For example, audio is more related to lip movements than torso motions. How to grasp the fine-grained appearance and dynamics of each portrait part remains unsolved; 2) Since the rays are uniformly sampled over the whole image, how to make the model pay more attention to small but important regions like mouth is challenging.

\noindent\textbf{Implicit Portrait Parsing Branch.} Our solution to the first problem is to add a parsing branch. Since the portrait parts of the same semantic category share similar motion patterns and texture information, it will be beneficial for the appearance and geometry learning in NeRF, which is also proven in recent implicit representation studies~\cite{Zhi:etal:ICCV2021, yariv2020multiview, yang2021objectnerf, yariv2021volume}. As shown in Fig.~\ref{fig:framework}, we extend the original NeRF with an additional parsing branch that predicts the semantic information. Note that since a certain 3D coordinate's semantic label is view-invariant, the parsing branch does not condition on view direction $\mathbf{d}$. Specifically, suppose there are totally $K$ semantic categories, the parsing branch maps the 3D spatial coordinate $\mathbf{x}$ to semantic logits $\mathbf{s(x)}$ over $K$ classes, which is further conditioned on audio $\mathbf{a}$. Hence the expected semantic logits $\hat{S}(\mathbf{r})$ along the ray $\mathbf{r}(v)$ with near and far bounds $v_n$ and $v_f$ can be calculated as:
\begin{align} 
    \label{eq:semantic}
\hat{S}(\mathbf{r}) &= \int_{v_n}^{v_f}T(v)\sigma(\mathbf{r}(v), \mathbf{a})\mathbf{s}(\mathbf{r}(v), \mathbf{a})dv,\\
    \label{eq:semantic2}
\text{where} \quad T(v) &= \exp(-\int_{v_n}^v\sigma(\mathbf{r}(u), \mathbf{a})du).
\end{align}

Such semantic awareness can naturally distinguish each part over the whole image, thus figuring out different associations between audio and different portrait regions.

\noindent\textbf{Dynamic Ray Sampling Strategy.} To generate delicate facial images with lip-synced results, we have to care for each portrait part, especially those small but crucial regions. Original NeRF \textbf{uniformly} samples rays on the image plane~\cite{mildenhall2020nerf}. Such an unconstrained ray sampling process focuses on big regions (\emph{e.g.}, background and cheek) yet ignores small regions (\emph{e.g.}, lip and teeth) that are important for fine-grained results. Therefore, we use semantic information to guide the ray sampling process dynamically. In particular, we denote all the points that are sampled on the image as $\Omega = \bigcup_{i=1}^K \Omega_i$, where $K$ is the total number of semantic categories in parsing map and $\Omega_i$ is the set of points that are sampled on the $i$-th semantic class. During the training stage, we calculate the average loss of each category $\mathcal{L}_i$ for the previous epoch (the sum of semantic loss and RGB loss, which will be introduced in Sec.~\ref{sec:3.4}), and then dynamically sample rays across $K$ categories by:
\begin{align} 
    \label{eq:dynamic}
    N_{\Omega_i} = \frac{\mathcal{L}_i}{\sum_{i=1}^K \mathcal{L}_i} \cdot N_s,
\end{align}
where $N_{\Omega_i}$ denotes the number of rays distributed to the $i$-th category and $N_s$ is the total number of sampled rays. We identify two benefits for such design: 1) The average loss of a semantic category is area-agnostic. Thus the learning process will equally sample those small-area regions; 2) Some image parts are comparatively easier to learn. For example, the texture of eye is more complicated than that of background. This leads to lower loss of background category and dynamically drives the implicit function to pay more attention to hard-to-learn regions. Our experiment further shows that this design can accelerate training as well.

\noindent\textbf{Structured 3D Information.} 3D cues are crucial for NeRF to grasp better spatial geometry information as proved in~\cite{xu2021generative, deng2021depth, wang2021neus, zhang2021ners}. In our framework, we identify that the awareness of \emph{rough} 3D facial information can serve as guidance for face semantic and geometry learning. Concretely, a 3D facial model is built with \emph{mean} expression parameters. We take inspiration from~\cite{peng2021neural, peng2020convolutional, yan2018second} to anchor a set of latent codes to the vertices of 3DMM model and diffuse to 3D space with SparseConvNet~\cite{graham20183d} to extract latent code volume. We query the latent code $\mathbf{f} \in \mathbb{R}^{88}$ at each point by trilinear interpolation\footnote{Please refer to the supplementary material for more details.} similar to Peng~\etal~\cite{peng2021neural}. Such structured 3D information could enhance semantic learning by giving similar features to the same semantic class while discriminative features among different semantic categories.
Till now, we can update the \textbf{basic} setting in Eq.~\ref{eq:basic} to \textbf{semantic-aware} implicit function with parameters $\Theta$:
\begin{equation} 
    \label{eq:implicit-semantic}
F^{\text{semantic}}_{\Theta} : (\mathbf{x}, \mathbf{d}, \mathbf{a}, \mathbf{f}) \rightarrow (\mathbf{c}, \sigma, \mathbf{s}).
\end{equation}

\subsection{Torso Deformation Module}
\label{sec:3.3}
As mentioned in Sec.~\ref{sec:3.1}, the estimated head pose serves as camera pose. However, such straightforward treatment ignores the fact that head and torso motions are inconsistent. To tackle this problem, we design a Torso Deformation module to stabilize the large-scale non-rigid torso motions.

\noindent\textbf{Torso Deformation Implicit Function.} Concretely, an implicit function is optimized to estimate the deformation field of $\Delta\mathbf{x} = (\Delta x, \Delta y, \Delta z)$ at a specific time instant $t$. 
Based on the observation that torso pose changes slightly and is weakly related to speech audio, the head pose $\mathbf{p}_\mathrm{h}(t)$ at time $t$ and a canonical pose $\mathbf{p}_\mathrm{c}$ are further given as references to learn the displacement $\Delta\mathbf{x}$, while audio feature $\mathbf{a}$ does not serve as input. 
Note that for convenience, the canonical pose $\mathbf{p}_\mathrm{c}$ is set as the head pose of the first frame, thus the displacement $\Delta\mathbf{x} = 0$ when $t = 0$. The implicit function for torso deformation with parameters $\Phi$ is formulated as:
\begin{equation} 
    \label{eq:implicit-deformation}
F^{\text{deform}}_{\Phi} : (\mathbf{x}, t, \mathbf{p}_\mathrm{h}(t), \mathbf{p}_\mathrm{c}) \rightarrow \Delta\mathbf{x}.
\end{equation}

Notably, although such deformation is added to the \emph{whole image}, we empirically find that \emph{only the torso part} tends to be deformed, while the facial dynamics are naturally modeled by semantic-aware implicit function in Eq.~\ref{eq:implicit-semantic}. Such disentanglement will be further analyzed in Sec.~\ref{sec:4.5}.

\noindent\textbf{Overall Implicit Function.} Combine the semantic-aware implicit function with our proposed Torso Deformation module, we can model the overall implicit function as:
\begin{align} 
    \label{eq:overall}
    &F^{\text{overall}}_{\Theta} : (\mathbf{x} + \Delta\mathbf{x}, \mathbf{d}, \mathbf{a}, \mathbf{f}) \rightarrow (\mathbf{c}, \sigma, \mathbf{s}), \notag \\
    \text{where} \quad &\Delta\mathbf{x} = F^{\text{deform}}_{\Phi}(\mathbf{x}, t, \mathbf{p}_\mathrm{h}(t), \mathbf{p}_\mathrm{c}).
\end{align}

\subsection{Volume Rendering and Network Training}
\label{sec:3.4}
\noindent\textbf{Volume Rendering with Deformation.} Since the Torso Deformation module is proposed to compensate for non-rigid torso motions, we accordingly adapt the NeRF's original volume rendering formulas for color and semantic distribution in Eq.~\ref{eq:nerf}, Eq.~\ref{eq:semantic} and Eq.~\ref{eq:semantic2}. Consider a certain 3D point $\mathbf{x}(v) = \mathbf{o} + v\mathbf{d}$ located on the ray emitted from center $\mathbf{o}$ on view direction $\mathbf{d}$, its warped coordinate at time $t$ with head pose $\mathbf{p}_\mathrm{h}(t)$ and canonical pose $\mathbf{p}_\mathrm{c}$ is computed as:
\begin{equation} 
\mathbf{x}^{\prime}(v, t) = \mathbf{x}(v) + F^{\text{deform}}_{\Phi}(\mathbf{x}(v), t, \mathbf{p}_\mathrm{h}(t), \mathbf{p}_\mathrm{c}).
\end{equation}

With the deformed 3D coordinate $\mathbf{x}^{\prime}(v, t)$ along the modified ray path $\mathbf{r}^{\prime}(v, t)$, we can calculate the expected color $\hat{C}(\mathbf{r}^{\prime}(v), t)$ and semantic logits $\hat{S}(\mathbf{r}^{\prime}(v), t)$ with near and far bounds $v_n$ and $v_f$ under \textbf{semantic-aware} setting as:
\begin{align}
    \label{eq:overall_render}
    &\hat{C}(\mathbf{r}^{\prime}) = \int_{v_n}^{v_f}T^{\prime}(v, t)\sigma(\mathbf{r}^{\prime}(v, t), \mathbf{a}, \mathbf{f})\mathbf{c}(\mathbf{r}^{\prime}(v, t), \mathbf{d}, \mathbf{a}, \mathbf{f})dv, \notag\\
    &\hat{S}(\mathbf{r}^{\prime}) = \int_{v_n}^{v_f}T^{\prime}(v, t)\sigma(\mathbf{r}^{\prime}(v, t), \mathbf{a}, \mathbf{f})\mathbf{s}(\mathbf{r}^{\prime}(v, t), \mathbf{a}, \mathbf{f})dv, \notag \\
  &\text{and} \quad T^{\prime}(v, t)= \exp(-\int_{v_n}^v\sigma(\mathbf{r}^{\prime}(u, t), \mathbf{a}, \mathbf{f})du),
\end{align}
where $T^{\prime}(v, t)$ is the accumulated transmittance along the ray path $\mathbf{r}^{\prime}(v, t)$ from $v_n$ to $v$. Note that the estimated semantic logits $\hat{S}(\mathbf{r}^{\prime})$ are subsequently transformed into multi-class distribution $p(\mathbf{r}^{\prime})$ through softmax operation. 

\noindent\textbf{Network Training.} Similar to NeRF~\cite{mildenhall2020nerf} that simultaneously optimizes coarse and fine models with hierarchical volume rendering, we train the network with following photometric loss $\mathcal{L}_\mathrm{p}$ and semantic loss $\mathcal{L}_\mathrm{s}$:
\begin{align}
    \label{eq:overall_loss}
    &\mathcal{L}_\mathrm{p} =\sum_{\mathbf{r}^{\prime}\in \mathcal{R}^{\prime}}\left[{\left\lVert \hat{C}_c(\mathbf{r}^{\prime}) - C(\mathbf{r}^{\prime}) \right\rVert}_2^2+{\left\lVert \hat{C}_f(\mathbf{r}^{\prime}) - C(\mathbf{r}^{\prime}) \right\rVert}_2^2 \right], \notag\\
    &\mathcal{L}_\mathrm{s} =- \sum_{\mathbf{r}^{\prime}\in \mathcal{R}^{\prime}}\left[
\sum_{k=1}^{K} p^{k}(\mathbf{r}^{\prime})\log \hat{p}_{c}^{k}(\mathbf{r}^{\prime})+
\sum_{k=1}^{K} p^{k}(\mathbf{r}^{\prime})\log \hat{p}_{f}^{k}(\mathbf{r}^{\prime})
\right],
\end{align}
where $\mathcal{R}^{\prime}$ is the set of \emph{deformed} camera rays passing through image
pixels; $C(\mathbf{r}^{\prime})$, $\hat{C}_c(\mathbf{r}^{\prime})$ and $\hat{C}_f(\mathbf{r}^{\prime})$ denote the ground-truth, coarse volume predicted and fine volume predicted pixel color for the deformed ray $\mathbf{r}^{\prime}$, respectively; and $p^{k}(\mathbf{r}^{\prime})$, $\hat{p}_{c}^{k}(\mathbf{r}^{\prime})$ and $\hat{p}_{f}^{k}(\mathbf{r}^{\prime})$ denote the ground-truth, coarse volume predicted and fine volume predicted multi-class semantic distribution for the deformed ray $\mathbf{r}^{\prime}$, respectively. The overall learning objective for the framework is:
\begin{align}
    \label{eq:total_loss}
    \mathcal{L} = \mathcal{L}_\mathrm{p} + \lambda \mathcal{L}_\mathrm{s},
\end{align}
where $\lambda$ is the weight balancing coefficient. At the training stage, the network parameters $\Theta$ and $\Phi$ of the implicit functions in Eq.~\ref{eq:overall} are updated based on above loss function.
\section{Experiments}
\label{sec:4}
\subsection{Dataset and Preprocessing}
\label{sec:4.1}
\noindent\textbf{Dataset Collection.} Our method targets to synthesize audio-driven facial images. Hence a certain person's speaking portrait video with audio track is needed. Unlike previous studies that demand large-corpus data or hours-long videos, we can achieve high-fidelity results with short videos of merely a few minutes. In particular, we extend the \emph{publicly-released} video set of Guo \etal~\cite{guo2021adnerf} and obtain videos of average length 6,750 frames in 25 fps.\textsuperscript{\ref{fotnt}}

\noindent\textbf{Training Data Preprocessing.} We follow the basic setting~\cite{guo2021adnerf} of audio-driven video portrait generation to preprocess training data: (1) For the speech audio, it is first processed by a pretrained DeepSpeech~\cite{amodei2016deep} model. Then a 1D convolutional network with self-attention mechanism is adopted~\cite{thies2020neural, guo2021adnerf} for smooth feature learning. The extracted audio feature $\mathbf{a} \in \mathbb{R}^{64}$ is fed into the implicit function in Eq.~\ref{eq:overall}. (2) For the video frames, they are cropped and resized to $450 \times 450$ to make talking portrait in the center. An off-the-shelf method~\cite{lee2020maskgan} is leveraged to obtain parsing maps of total 11 semantic classes. The background image and head pose are estimated in a similar way to Guo \etal~\cite{guo2021adnerf}. Note that the estimated head pose $\mathbf{p}_\mathrm{h}(t)$ at time $t$ is treated as camera pose and the canonical pose $\mathbf{p}_\mathrm{c}$ is set as the starting frame's head pose, \emph{i.e.}, $\mathbf{p}_\mathrm{c} = \mathbf{p}_\mathrm{h}(0)$ in Eq.~\ref{eq:overall}.

\setlength{\tabcolsep}{9pt}
\begin{table*}
  \centering
  \begin{tabular}{lcccccccccc}
    \toprule
     & \multicolumn{4}{c}{Testset A} & \multicolumn{2}{c}{Testset B~\cite{thies2020neural}} &  \multicolumn{2}{c}{Testset C~\cite{suwajanakorn2017synthesizing}} \\
    \cmidrule(r){2-5} \cmidrule(r){6-7} \cmidrule(r){8-9}
    Methods & PSNR $\uparrow$ & SSIM $\uparrow$ & LMD $\downarrow$ & Sync $\uparrow$ & LMD $\downarrow$ & Sync $\uparrow$ & LMD $\downarrow$ & Sync $\uparrow$\\
    Ground Truth & N/A & 1.000 & 0 & 6.632 & 0 & 5.973 & 0 & 6.204\\
    \midrule
     ATVG~\cite{chen2019hierarchical} & 24.125 & 0.725 & 5.261 & 4.708 & 5.074 & 6.208 & 5.869 & 4.419\\
     Wav2Lip~\cite{prajwal2020lip} & 26.667 & 0.793 & 5.811 & \textbf{6.952} & 4.893 & \textbf{6.980} & 5.740 & \textbf{6.806}\\
     MakeitTalk~\cite{zhou2020makelttalk} & 25.522 & 0.704 & 7.238 & 3.873 & 6.704 & 4.105 & 6.512 & 3.925\\
     PC-AVS~\cite{zhou2021pose} & 25.712 & 0.756 & 5.406 & 5.834 & 5.247 & 6.113 & 5.771 & 5.983\\
     NVP~\cite{thies2020neural} & - & - & - & - & 5.072 & 5.689 & - & -\\
     SynObama~\cite{suwajanakorn2017synthesizing} & - & - & - & - & - & - & 5.485 & 5.938\\
     AD-NeRF~\cite{guo2021adnerf} & 29.814 & 0.844 & 5.183 & 6.092 & 5.119 & 5.613 & 5.392 & 6.012\\
     \midrule 
     \textbf{SSP-NeRF (Ours)} & \textbf{32.649} & \textbf{0.868} & \textbf{4.934} & 6.438 & \textbf{4.892} & 5.886 & \textbf{5.208} & 6.186\\
    \bottomrule
  \end{tabular}
  \caption{\textbf{The quantitative results of \emph{cropped setting} on Testset A, B~\cite{thies2020neural} and C~\cite{suwajanakorn2017synthesizing}.} We compare the proposed Semantic-aware Speaking Portrait NeRF \textbf{(SSP-NeRF)} against recent SOTA methods~\cite{chen2019hierarchical, prajwal2020lip, zhou2020makelttalk, zhou2021pose, thies2020neural, suwajanakorn2017synthesizing, guo2021adnerf} and ground truth under four metrics. For LMD the lower the better, and the higher the better for other metrics. Note that the detailed comparison settings are elaborated in Sec.~\ref{sec:4.3}.}
  \label{tbl:crop}
  \vspace{-2mm}
\end{table*}

\subsection{Experimental Settings}
\label{exp-setting}
\noindent\textbf{Comparison Baselines.}
We compare our method with recent representative works: (1) \textbf{ATVG}~\cite{chen2019hierarchical}, which uses 2D landmark to guide facial image synthesis; (2) \textbf{Wav2Lip}~\cite{prajwal2020lip} that achieves state-of-the-art lip-sync performance by pretraining a lip-sync expert; (3) \textbf{MakeitTalk}~\cite{zhou2020makelttalk}, a representative 3D landmark-based approach; (4) \textbf{PC-AVS}~\cite{zhou2021pose} which generates pose-controllable talking face by modularized audio-visual representation; (5) \textbf{NVP}~\cite{thies2020neural} that first infers expression parameters from audio, then generates images with a neural renderer; (6) \textbf{SynObama}~\cite{suwajanakorn2017synthesizing} which learns mouth shape changes for facial image warping; (7) \textbf{AD-NeRF}~\cite{guo2021adnerf}, which is the first work that uses implicit representation of NeRF to achieve arbitrary-size talking head synthesis. In particular, we also show the evaluations directly on the \textbf{Ground Truth} for a clearer comparison.

\setlength{\tabcolsep}{4pt}
\begin{table}
  \centering
  \small
  \begin{tabular}{lccccc}
    \toprule
    & \multicolumn{4}{c}{Testset A ($450\times450$)} & \\
    \cmidrule(r){2-5}
    Methods & PSNR & SSIM & LMD & Sync & \# of params\\
    GT & N/A & 1.000 & 0 & 5.291 & -\\
    \midrule
    AD-NeRF & 29.186 & 0.827 & 4.892 & 4.237 & 2.69M\\
    \textbf{Ours} & \textbf{32.785} & \textbf{0.876} & \textbf{4.495} & \textbf{4.993} & \textbf{1.10M}\\
    \bottomrule
  \end{tabular}
  \caption{\textbf{The quantitative results of \emph{full resolution setting} on Testset A.} We compare our method with AD-NeRF~\cite{guo2021adnerf} that also generates whole portrait with full resolution of $450\times450$. We evaluate image quality and lip-sync accuracy of synthesized results. The number of parameters for each model is shown in table.}
  \label{tbl:full}
\end{table}

\noindent\textbf{Implementation Details.}\footnote{Please refer to supplementary material for more details\label{fotnt}.} The $F^{\text{semantic}}_{\Theta}$ and $F^{\text{deform}}_{\Phi}$ together with their associated fine models all consist of simple 8-layers MLPs with hidden size of 128 and ReLU activations. Following NeRF~\cite{mildenhall2020nerf}, positional encoding is applied to each 3D coordinate $\mathbf{x}$, view direction $\mathbf{d}$ and time instant $t$ to map the input into higher dimensional space for better learning. The positional encoder is formulated as: $\gamma(q) = <(\sin(2^l\pi q), \cos(2^l\pi q))>^L_0$, where we use $L=10$ for $\mathbf{x}$, and $L=4$ for $\mathbf{d}$ and $t$. For the parsing maps, we use $K=11$ categories for semantic guidance, including cheek, eye, eyebrow, ear, nose, teeth, lip, neck, torso, hair and background. The structured 3D feature extractor is borrowed from~\cite{peng2021neural} that processes feature volume with 3D sparse convolutions and outputs latent code with $2\times$, $4\times$, $8\times$, $16\times$ downsampled sizes. The semantic weight $\lambda$ is empirically set to $0.04$. The model is trained with $450\times450$ images during $400k$ iterations with a batch size of $N_s = 1024$ rays. The framework is implemented in PyTorch~\cite{paszke2019pytorch} and trained with Adam optimizer~\cite{adam} of learning rate $5e-4$ on a single Tesla V100 GPU for 36 hours.

\subsection{Quantitative Evaluation}
\label{sec:4.3}
\noindent\textbf{Evaluation Metrics.}
We employ evaluation metrics that have been previously used in talking face generation. We adopt \textbf{PSNR} and \textbf{SSIM}~\cite{wang2004image} to evaluate the image quality of generated results; Landmark Distance (\textbf{LMD})~\cite{chen2018lip} and \textbf{SyncNet Confidence}~\cite{chung2016lip, Chung16a} to account for the accuracy of mouth shapes and lip sync. Note that the landmarks are detected from synthesized images for the computation of LMD metric. Other metrics such as \textbf{CSIM}~\cite{chen2020talking, zakharov2019few} for measuring identity preserving and \textbf{CPBD}~\cite{narvekar2009no} for measuring result sharpness are shown in supplementary material.

\begin{figure*}[t!]
    \centering
    \includegraphics[width=0.9\linewidth]{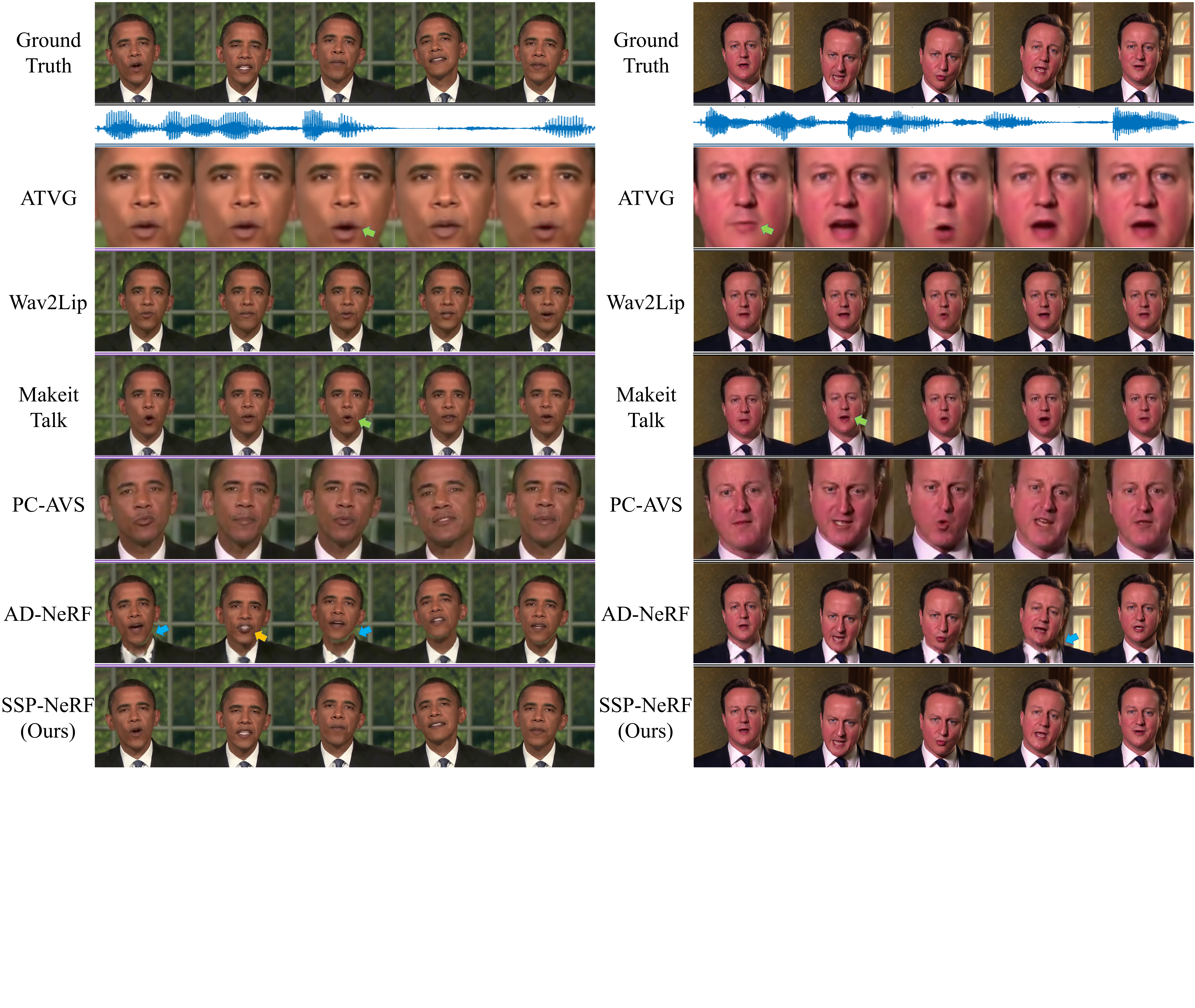}
    \vspace{-2mm}
    \caption{\textbf{The comparison of generated key frame results on Testset A.} We show the synthesized talking heads of ground truth, baseline methods~\cite{chen2019hierarchical, prajwal2020lip, zhou2020makelttalk, zhou2021pose, guo2021adnerf} and ours. Please \textbf{zoom in for better visualization}. More qualitative comparisons can be found in demo video.}
    \label{vis}
    \vspace{-2mm}
\end{figure*}

\setlength{\tabcolsep}{6pt}
\begin{table*}
  \centering
  \small
  \begin{tabular}{ccccccccc}
    \toprule
    Methods &
    ATVG & 
    Wav2Lip & MakeitTalk & 
    PC-AVS &
    NVP &
    SynObama &
    AD-NeRF & 
    \textbf{SSP-NeRF (Ours)} \\
    \midrule
    Lip-sync Accuracy & 3.02 & 4.23 & 2.89 & 4.05 & \textbf{4.26} & 4.21 & 4.16 &\textbf{4.26}\\
    Video Realness & 1.63 & 2.86 & 2.45 & 3.83 & 3.89 & 3.64 & 4.09 &\textbf{4.28}\\
    Image Quality & 1.72 & 2.42 & 2.78 & 2.36 & 4.02 & 3.49 & 4.18 &\textbf{4.43}\\
    \bottomrule
  \end{tabular}
  \vspace{-3mm}
  \caption{\textbf{User study results on the generation quality of audio-driven portrait.} The rating is of scale 1-5, with the larger the better. We compare the lip-sync accuracy, video realness and image quality of baseline methods~\cite{chen2019hierarchical, prajwal2020lip, zhou2020makelttalk, zhou2021pose, guo2021adnerf, thies2020neural, suwajanakorn2017synthesizing} and our proposed \textbf{SSP-NeRF}.}
  \label{table:userstudy}
  \vspace{-1mm}
\end{table*}

\noindent\textbf{Comparison Settings.}
The reconstruction/model-based methods require large-corpus training data or long videos, hence we directly inference with their publicly-released best models. Note that all baseline methods except for~\cite{guo2021adnerf} fail to generate the whole portrait with full resolution, we divide our comparisons into two settings: 1) The \emph{cropped setting} in Table~\ref{tbl:crop}, where we crop the generated facial image with same region and resize into same size for fair evaluation metric comparison. 2) The \emph{full resolution setting} in Table~\ref{tbl:full}, where we compare with AD-NeRF~\cite{guo2021adnerf} that could also synthesize the whole portrait with full resolution of $450\times450$.

In the first setting, since NVP~\cite{thies2020neural} and SynObama~\cite{suwajanakorn2017synthesizing} do not provide pretrained models, we conduct comparisons on three datasets: (1) \textbf{Testset A}, the collected dataset mentioned in Sec.~\ref{sec:4.1}; (2) \textbf{Testset B}, where we extract speech audio from the demo of NVP to drive other baselines; (3) \textbf{Testset C}, where the audio from SynObama's demo is used for animation. Note that the metrics for measuring image quality (PSNR and SSIM) are not evaluated on Testset B and C due to the low image quality of original videos. In the second setting, the experiment is only conducted on Testset A for high-resolution comparison. We further compare the number of model parameters against AD-NeRF~\cite{guo2021adnerf} to show the efficiency of our proposed approach.

\noindent\textbf{Evaluation Results.}
The results of the \emph{cropped setting} and \emph{full resolution setting} are shown in Table~\ref{tbl:crop} and Table~\ref{tbl:full}, respectively. It can be seen that the proposed \textbf{SSP-NeRF} achieves the best evaluation results in most metrics: \textbf{(1)} In the cropped setting, we synthesize fine-grained facial images with detailed local appearance and dynamics of each portrait part. Note that Wav2Lip~\cite{prajwal2020lip} uses SyncNet~\cite{chung2016lip, Chung16a} for pretraining, which makes their results on SyncNet Confidence even better than the ground truth. Our performance on the LMD metric is the best, and the SyncNet Confidence of our model is close to the ground truth on all three datasets, showing that we can generate accurate lip-sync video portraits. \textbf{(2)} In the full resolution setting, the human face as well as torso part is evaluated. Different from AD-NeRF's separated rendering pipeline, our design of Torso Deformation module facilitates steady results. The statistics on both model's parameter number are shown in Table~\ref{tbl:full}. Notably, our method is trained with $400k \times 1024$ sampled rays, while AD-NeRF~\cite{guo2021adnerf} uses $400k \times 2048$ rays for each model. Hence we generate portraits of \textit{better} image quality and \textit{better} lip-synchronization in a \textit{more compact} model with \textit{fewer} iterations, proving the effectiveness and efficiency of \textbf{SSP-NeRF}.\textsuperscript{\ref{fotnt}}

\subsection{Qualitative Evaluation}
\label{sec:4.4}
To compare the generated results of each method, we show the key frames of two clips in Fig.~\ref{vis}. The figure shows that our method synthesizes more lip-synced video portraits of higher image quality. In particular, ATVG~\cite{chen2019hierarchical} and MakeitTalk~\cite{zhou2020makelttalk} rely on precise facial landmarks, which leads to inaccurate mouth shapes (green arrow); Wav2Lip~\cite{prajwal2020lip} creates static talking heads; PC-AVS~\cite{zhou2021pose} fails to preserve the speaker's identity, making generated results unrealistic. Moreover, all the image reconstruction-based methods~\cite{prajwal2020lip, zhou2021pose} or model-based methods~\cite{chen2019hierarchical, zhou2020makelttalk} fail to synthesize the whole portrait of high-fidelity simultaneously. Although AD-NeRF~\cite{guo2021adnerf} manages to create full-resolution results, the separated rendering pipeline with uniform ray sampling leads to head-torso separation (as highlighted by blue arrows) and blurry results (orange arrow). 

\noindent\textbf{User Study.}\textsuperscript{\ref{fotnt}} Since subjective evaluation can reflect the quality of audio-driven portrait, a user study is further conducted. Specifically, we sample 30 audio clips from Testset A, B and C for all methods to generate results, and then involve 18 participants for user study. The Mean Opinion Scores rating protocol is adopted for evaluation, which requires the participants to rate three aspects of generated speaking portraits: (1) \textit{Lip-sync Accuracy}; (2) \textit{Video Realness}; (3) \textit{Image Quality}. The rating is based on a scale of 1 to 5, with 5 being the maximum and 1 being the minimum.

The results are shown in Table~\ref{table:userstudy}. Since NeRF enables full-resolution whole portrait generation with expressive pose, both AD-NeRF~\cite{guo2021adnerf} and our method score comparatively high on \textit{Image Quality} and \textit{Video Realness}. Besides, the users prefer our generated speaking portraits to AD-NeRF's~\cite{guo2021adnerf} due to the fine-grained local rendering and stable torso motions provided by our framework design. Although PC-AVS~\cite{zhou2021pose} also creates pose-controllable talking faces, the inaccuracy of implicit pose code extraction weakens their realness. Note that Wav2Lip~\cite{prajwal2020lip}, NVP~\cite{thies2020neural} and SynObama~\cite{suwajanakorn2017synthesizing} achieve competitive scores on \textit{Lip-sync Accuracy}. However, they rely on large corpus or long training videos, while we merely take a short video as input, showing the efficacy of our method. To further measure the disagreement on scoring among the participants, the Fleiss's-Kappa\footnote {\href{https://en.wikipedia.org/wiki/Fleiss\%27\_kappa}{https://en.wikipedia.org/wiki/Fleiss\%27\_kappa}} statistic is calculated on 18 participants' ratings. The Fleiss-Kappa value is $0.816$, which can be interpreted as ``almost perfect agreement''.

\begin{table}
  \centering
  \begin{tabular}{lcccc}
    \toprule
    Methods & PSNR $\uparrow$ & SSIM $\uparrow$ & LMD $\downarrow$ & Sync $\uparrow$\\
    \midrule
    w/o $F^{\text{deform}}_{\Phi}$ & 27.472 & 0.791 & 4.635 & 4.744\\
    deform by $\mathbf{a}$ & 28.013 & 0.802 & 5.329 & 3.871\\
    \textbf{SSP-NeRF} & \textbf{32.785} & \textbf{0.876} & \textbf{4.495} & \textbf{4.993}\\
    \bottomrule
  \end{tabular}
  \vspace{-2mm}
  \caption{\textbf{Ablation study results of Torso Deformation module.}}
  \label{tbl:ablation1}
  \vspace{-1mm}
\end{table}

\begin{figure}[t!]
    \centering
    \includegraphics[width=\linewidth]{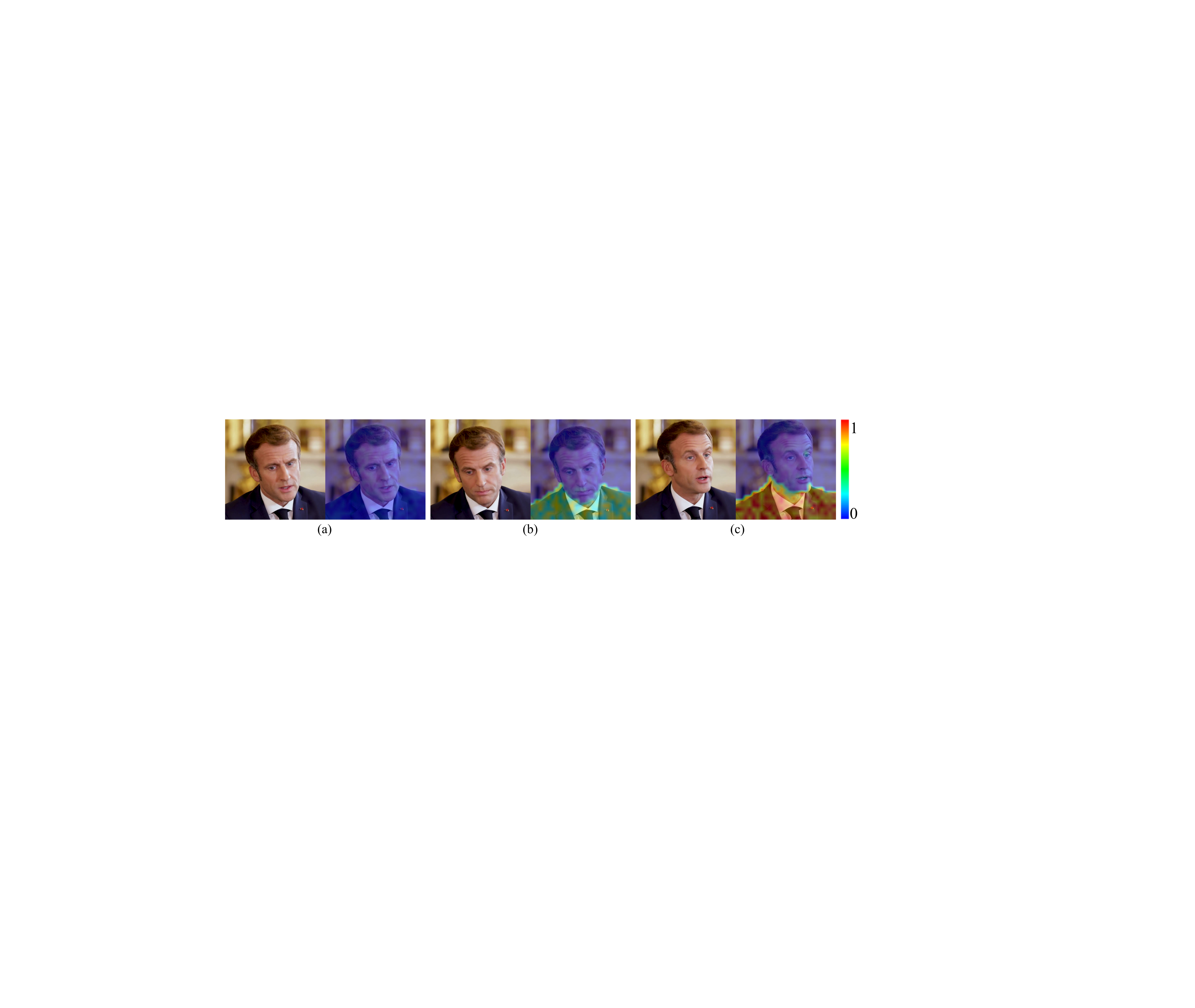}
    \vspace{-6mm}
    \caption{\textbf{The visualized deformation heatmap.} From left to right, we show the predicted displacements over the whole portrait image under three cases of small, medium and large pose. We can observe that: 1) The deformations are mostly on the torso region; 2) The larger pose is, the more displacements it will learn.}
    \label{deform}
    \vspace{-2mm}
\end{figure}

\subsection{Ablation Study}
\label{sec:4.5}
In this section, we present ablation study on the Testset A in terms of two key modules proposed in our framework.

\noindent\textbf{Torso Deformation Module.} We conduct ablation experiments under two settings: (1) w/o $F^{\text{deform}}_{\Phi}$, where we directly synthesize the whole portrait without deforming 3D coordinates. The results are shown in Table~\ref{tbl:ablation1} (line1), where the ill-posed rendering leads to blurry torso with low image quality. To further investigate the efficacy of Torso Deformation module, we visualize the heatmap of learned displacements in Fig.~\ref{deform}. Since audio feature is not input to the deformation implicit function, it tends to warp the weakly audio-related torso part, while the strongly audio-related mouth movements are mostly modeled by $F^{\text{semantic}}_{\Theta}$. The marginal drop in lip-sync metrics also suggests that the deformation module majorly takes effect on the torso part.

Another ablation setting is: (2) deform by $\mathbf{a}$, where the audio input $\mathbf{a}$ is fed to $F^{\text{deform}}_{\Phi}$ rather than $F^{\text{semantic}}_{\Theta}$, \emph{i.e.}, the audio feature $\mathbf{a}$, head pose $\mathbf{p}_\mathrm{h}(t)$ and canonical pose $\mathbf{p}_\mathrm{c}$ are leveraged to deform both the human face and torso part simultaneously. The lip-sync performance drops dramatically as shown in Table~\ref{tbl:ablation1} (line2). We guess the reason lies in distinct correlations between audio and different portrait parts. It is hard for deformation module to handle audio synchronization and portrait parts deformation at the same time.

\noindent\textbf{Semantic-aware Dynamic Ray Sampling Module.} The ablative experiments contain: (1) w/o semantic branch, which means the semantic supervision $\mathcal{L}_\mathrm{s}$ is not used; (2) w/o dynamic sample, where the rays are uniformly sampled over image plane; (3) w/o 3D information, which means the 3D feature $\mathbf{f}$ is eliminated. The results in Table~\ref{tbl:ablation2} verify that the semantic awareness enables the model to better grasp each part's appearance and geometry. The dynamic ray sampling further facilitates fine-grained results.

\begin{table}[t]
  \centering
  \small
  \begin{tabular}{lcccc}
    \toprule
    Methods & PSNR$\uparrow$ & SSIM$\uparrow$ & LMD$\downarrow$ & Sync$\uparrow$\\
    \midrule
    w/o semantic branch & 29.479 & 0.832 & 4.886 & 4.562\\
    w/o dynamic sample & 29.514 & 0.826 & 4.916 & 4.490\\
    w/o 3D information & 31.059 & 0.845 & 4.683 & 4.739\\
    \textbf{SSP-NeRF} & \textbf{32.785} & \textbf{0.876} & \textbf{4.495} & \textbf{4.993}\\
    \bottomrule
  \end{tabular}
  \caption{\textbf{Ablation study of Semantic-Aware Dynamic Ray Sampling Module.} The ablation settings are elaborated in Sec.~\ref{sec:4.5}.}
  \label{tbl:ablation2}
  \vspace{-2mm}
\end{table}
\section{Broader Impact}
\label{ethical}
\noindent\textbf{Ethical Consideration.} Animating realistic talking portrait has extensive applications like digital human and film-making. On the other hand, it could be misused for malicious purposes such as identity theft, deepfake generation, and media manipulation. Recent studies have shown promising results in detecting deepfakes~\cite{rossler2018faceforensics, rossler2019faceforensics++}. However, the lack of realistic data limits their performance. As part of our responsibility, we feel obliged to share our generated results with the deepfake detection community to improve the model's robustness. We believe that the proper use of this technique will enhance the healthy development of both machine learning research and digital entertainment.

\noindent\textbf{Limitation and Future Work.} Our proposed SSP-NeRF achieves audio-driven video portrait generation of high-fidelity. However, the method still has limitations. The speed of synthesizing images is slow due to the heavy computation of rendering high-quality images. We also observe that the language gap between training and driven audio makes the synthesized mouth look unnatural occasionally~\cite{guo2021adnerf}. We will address these issues in future work. 
\section{Conclusion}
\label{conclusion}
In this paper, we propose a novel framework Semantic-aware Speaking Portrait NeRF (\textbf{SSP-NeRF}) for audio-driven portrait generation. We introduce Semantic-Aware Dynamic Ray Sampling module to grasp the detailed appearance and the local dynamics of each portrait part without using accurate structural information. We then propose a Torso Deformation module to learn global torso motion and prevent head-torso separated results. Extensive experiments show that our approach can synthesize more realistic video portraits compared to the previous methods.

{\small
\bibliographystyle{ieee_fullname}
\bibliography{egbib}
}

\end{document}